\documentclass{article}
\usepackage{spconf,amsmath,graphicx}
\usepackage{amssymb}
\usepackage{booktabs}
\usepackage{pifont}
\usepackage{multirow}
\usepackage{cancel}
\usepackage{tikz}
\usetikzlibrary{positioning}
\usetikzlibrary{calc}
\usepackage{pgfplots, pgfplotstable}    
\pgfplotsset{compat=1.16} 
\usepackage{url}
\usepackage{algorithm,algpseudocode}
\DeclareMathSymbol{\shortminus}{\mathbin}{AMSa}{"39}
\newcommand{\argmax}{\mathop{\mathrm{argmax}}\limits}

\newcommand{\Sref}[1]{\S\ref{#1}}
\newcommand{\Fref}[1]{Figure~\ref{#1}}
\newcommand{\Tref}[1]{Table~\ref{#1}}

\newcommand{\Hquad}{\hspace{0.0em}} 
\newcommand\mypar[1]{\noindent\textbf{#1}\Hquad}

\usepackage{icomma}

\makeatletter
\def\blfootnote{\xdef\@thefnmark{}\@footnotetext}
\makeatother

\usepackage[
backend=biber,
style=ieee,
citestyle=numeric-comp,
maxbibnames=3,
maxcitenames=3,
doi=false,isbn=false,url=false,eprint=false
]{biblatex}

\defbibheading{bibliography}[\refname]{}

\title{Towards Zero-Shot Code-Switched Speech Recognition}

\name{
\begin{tabular}{c}
\it Brian Yan${}^1$, Matthew Wiesner${}^2$, Ond\v{r}ej Klejch${}^3$, Preethi Jyothi${}^4$, Shinji Watanabe${}^{1,2}$
\end{tabular}
}
\address{${}^1$Carnegie Mellon University, US, ${}^2$Johns Hopkins University, US, \\${}^3$University of Edinburgh, UK, ${}^4$Indian Institute of Technology Bombay, IN}

\begin{document}
\ninept
\maketitle
\begin{abstract}
In this work, we seek to build effective code-switched (CS) automatic speech recognition systems (ASR) under the zero-shot setting where no transcribed CS speech data is available for training.
Previously proposed frameworks which conditionally factorize the bilingual task into its constituent monolingual parts are a promising starting point for leveraging monolingual data efficiently.
However, these methods require the monolingual modules to perform \textit{language segmentation}.
That is, each monolingual module has to simultaneously detect CS points and transcribe speech segments of one language while ignoring those of other languages -- not a trivial task.
We propose to simplify each monolingual module by allowing them to transcribe all speech segments indiscriminately with a monolingual script (i.e. \textit{transliteration}).
This simple modification passes the responsibility of CS point detection to subsequent bilingual modules which determine the final output by considering multiple monolingual transliterations along with external language model information.
We apply this transliteration-based approach in an end-to-end differentiable neural network and demonstrate its efficacy for zero-shot CS ASR on Mandarin-English SEAME test sets.
\end{abstract}

\section{Introduction}

\label{sec:intro}

In order to build multilingual automatic speech recognition (ASR) systems that are robust to code-switching (CS), practitioners must tackle both the long-tail of possible language pairs \cite{lewis2009ethnologue} and the relative infrequency of intra-sententially CS examples within collected training corpora \cite{gamback2016comparing}.
Therefore, a preeminent challenge in the CS ASR field is to build effective systems under the zero-shot setting where no CS ASR training data is available. 
Recent advancements in multilingual speech recognition have demonstrated the impressive scale of cross-lingual sharing in neural network approaches \cite{watanabe2017language, li2019bytes, li2021scaling, Yan2021DifferentiableAG, lu2022language, bapna2022mslam, li2022asr2k, bai2022joint, zhou2022configurable, zhang2022streaming}, and these works have shown that jointly modeling ASR with language identity (LID) grants some intra-sentential CS ability \cite{zhou2022configurable, zhang2022streaming, seki2018end}.
However, most of these large scale models skew towards high-resourced languages \cite{li2022asr2k} and do not seek to directly optimize for intra-sentential CS ASR between particular language pairs.

A more promising direction towards zero-shot CS ASR can be found in prior works which seek to incorporate monolingual data directly to improve CS performance \cite{gonen2018language, li2019towards, shan2019component, taneja2019exploiting, shi2020asru, shah2020learning, lu2020bi, zhou2020multi, chuang2020training, dalmia2021transformer, zhang2021decoupling, liu2021code, Ali2021ArabicCS, diwan2021multilingual, deng2022summary}.
In particular, there are several works which achieve joint modeling of CS and monolingual ASR by conditionally factorizing the overall bilingual task into monolingual parts \cite{cond2022, tian2022lae, song2022language}.
By using label-to-frame synchronization, this \textit{conditionally factorized} approach can make a CS prediction given only the predictions of the monolingual parts \cite{cond2022} -- theoretically these conditionally factorized models can model CS ASR without any CS data, but this has not been previously confirmed.

In this work, we seek to build CS ASR systems under two zero-shot data conditions: 1) monolingual speech and CS text data are available, 2) only monolingual speech and text data are available.
In particular, we are interested in exploring the zero-shot capability of conditionally factorized joint CS and monolingual ASR models. 

We first re-formulate the initial monolingual stage of these conditionally factorized models in terms of their \textit{language segmentation} burden, showing that prior works expect each monolingual module to perform CS point detection and transcription in tandem.
Any errors in CS point detection are thus propagated downstream to the final bilingual stage which attempts to stitch multiple monolingual predictions into an output which may or may not be CS.
To improve model robustness towards zero-shot CS ASR, we propose an alternative formulation of the monolingual stage such that each module is an indiscriminate \textit{transliterator}, transcribing all speech using a monolingual script without any regard for potential CS points.
As a result we delay CS point detection until the final bilingual stage, allowing our models to condition this critical decision on multiple monolingual inputs and incorporate additional information from external language models.
Our transliteration-based method yielded $5$ absolute error-rate reduction in our zero-shot CS ASR experiments. 

\section{Background and Motivation}

\label{sec:background}

\begin{figure}
\centering
\includegraphics[width=\linewidth]{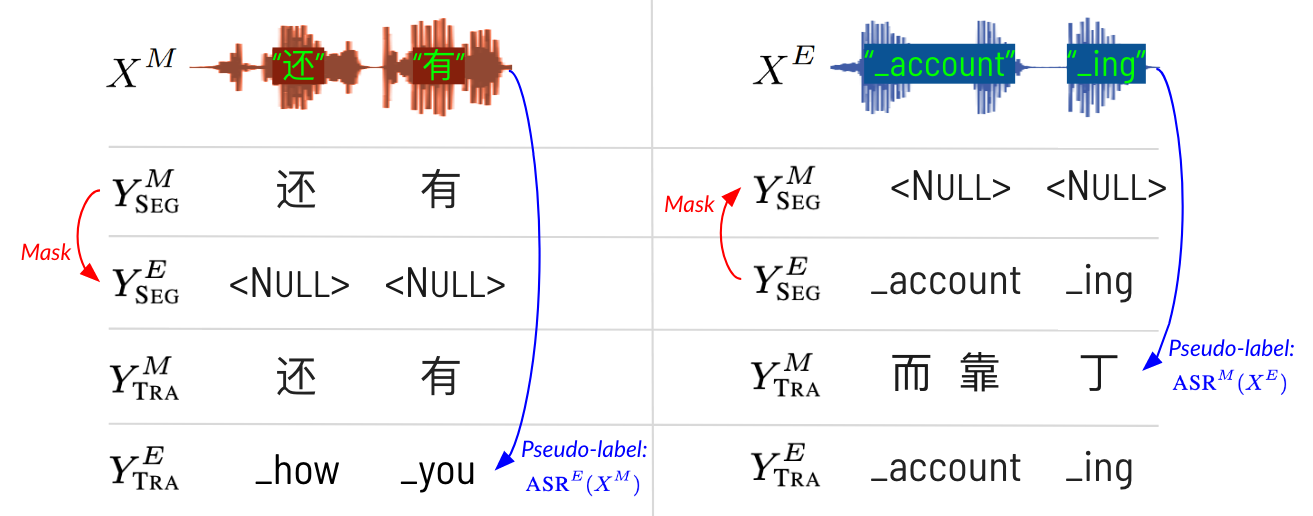}
\vspace{-5mm}
\caption{
Examples showing the difference between language segmentation targets $Y_{\textsc{Seg}}^{M/E}$ obtained via \textcolor{red}{masking} (\Sref{sec:lang_seg}) vs. transliteration targets $Y_{\textsc{Tra}}^{M/E}$ obtained via cross-lingual \textcolor{blue}{pseudo-labeling} (\Sref{sec:transliteration}).
}
\vspace{-2mm}
\label{fig:cs_target}
\end{figure}

In this section, we examine the language segmentation role of the monolingual modules in previously proposed conditionally factorized models \cite{cond2022}, motivating our transliteration-based approach (\Sref{sec:proposed}).

\subsection{Joint Modeling of Code-Switched and Monolingual ASR}
\label{sec:cond_background}

Let us take the Mandarin-English bilingual pair as an example for the following formulations.
Bilingual ASR, where speech may or may not be CS, is a sequence mapping from a $T$-length speech feature sequence, $X = \{ \mathbf{x}_t \in \mathbb{R}^D | t = 1, ..., T \}$, to an $L$-length label sequence, $Y = \{ y_l \in ( \mathcal{V}^M \cup \mathcal{V}^E ) | l = 1, ..., L \}$ consisting of Mandarin $\mathcal{V}^M$ and English $\mathcal{V}^E$.
The conditionally factorized framework \cite{cond2022} decomposes this bilingual task into three sub-tasks: 1) recognizing Mandarin, 2) recognizing English, and 3) composing recognized monolingual segments into a bilingual sequence. 
% which may or may not be intra-sententially CS. 

The basis of this approach is to model the label-to-frame alignments. For each $T$-length observation sequence $X$ and $L$-length bilingual label sequence $Y$ there are a number of possible $T$-length label-to-frame sequences $Z = \{ z_t \in \mathcal{V}^M \cup \mathcal{V}^E \cup \{\varnothing \} | t=1\ldots T \}$, where $\varnothing$ denotes a blank symbol as in Connectionist Temporal Classification (CTC) \cite{graves2006connectionist} or RNN-T \cite{rnnt_graves}.
Further consider that for each bilingual $Z$ there are two corresponding monolingual label-to-frame sequences $Z^M = \{ z^M_t \in \mathcal{V}^M \cup \{\varnothing \} | t=1\ldots T \}$ and $Z^E = \{ z^E_t \in \mathcal{V}^E \cup \{\varnothing \} | t=1\ldots T \}$.
The label posterior, $p(Y|X)$, can thus be represented in terms of bilingual, $p(Z|X)$, and monolingual, $p(Z^{M}|X)$ and $p(Z^{E}|X)$, label-to-frame posteriors as follows:

\begin{align}
    % p(Y | X) &= \sum_{Z \in \mathcal{Z}} p(Z | X) \label{eq:1} \\
    p(Y | X) &= \sum_{Z \in \mathcal{Z}} \sum_{Z^M \in \mathcal{Z}^M} \sum_{Z^E \in \mathcal{Z}^E} p(Z, Z^M, Z^E | X) \label{eq:exact}
\end{align}
where $\mathcal{Z}$ and $\mathcal{Z}^{M/E}$ denote sets of all possible bilingual and monolingual label-to-frame alignments for a given $Y$.
Eq.~\eqref{eq:exact} is the exact \textit{joint} bilingual and monolingual ASR likelihood which can be further factorized using independence assumptions to obtain the form:

\begin{align}
    p(Y | X) &\approx \underbrace{\sum_{Z} p(Z | Z^M, Z^E)}_{\text{Bilingual Posterior}} 
    \underbrace{\sum_{Z^M} p(Z^M|X) \sum_{Z^E} p(Z^E|X)}_{\text{Monolingual Posteriors}} 
    \label{eq:approx}
\end{align}
From Eq.~\eqref{eq:exact} to Eq.~\eqref{eq:approx}, the first assumption is that given $Z^M$ and $Z^E$, no other information from the observation $X$ is required to determine $Z$, allowing for conditional modeling of the bilingual posterior $p(Z | Z^M, Z^E, \cancel{X})$ given only monolingual information.
The second assumption is that given $X$, $Z^M$ and $Z^E$ are independent, allowing for separate modeling of monolingual posteriors $p(Z^M| \cancel{Z^E}, X)$ and $p(Z^E| \cancel{Z^M}, X)$.
Note we abbreviate this pair of separate monolingual modules as $p(Z^{M/E}|X)$ in future sections.

\subsection{Modeling $p(Z^{M/E}|X)$ with Language Segmentation}
\label{sec:lang_seg}

What should be the behavior of the monolingual Mandarin module $p(Z^M|X)$ when encountering a segment of English speech and vice versa?
Monolingual modules in prior works \cite{cond2022, tian2022lae, song2022language} determine each label-to-frame alignment $z_t^{M/E}$ by first determining the language identity of each speech frame $\operatorname{LID}(\mathbf{x}_t)$ \cite{liu2021end}. If the speech frame $\mathbf{x}_t$ is from a foreign language then the module will ignore it by emitting a special $\textless \textsc{Null} \textgreater$ token, otherwise it will transcribe using its monolingual vocabulary. This monolingual \textit{language segmentation} decision is defined as follows (shown for Mandarin):
\begin{align}
    z_t^M = 
    \begin{cases}
        \argmax_{m\in  \mathcal{V}^M \cup \{ \varnothing \}} p(z_t^M=m|X, z_{1:t-1}^M) & \text{if $\operatorname{LID}(\mathbf{x}_t)$ is $M$} \\
        \argmax_{m \in \{ \text{\textless}\textsc{Null}\text{\textgreater}, \varnothing \}} p(z_t^M=m|X, z_{1:t-1}^M) & \text{if $\operatorname{LID}(\mathbf{x}_t)$ is $E$} \label{eq:seg}
    \end{cases}
\end{align}
Note that the frame-wise $\operatorname{LID}(\mathbf{x}_t)$ is not a separate module, but rather an implicit decision within the posterior maximization over the $\textless \textsc{Null} \textgreater$ augmented monolingual label-to-frame alignments $Z^{M/E} = \{ z^{M/E}_t \in \mathcal{V}^{M/E} \cup \{\varnothing, \text{\textless} \textsc{Null} \text{\textgreater} \} | t=1\ldots T \}$.
This language segmentation behavior is learned by optimizing likelihoods of $\textless \textsc{Null} \textgreater$ masked label targets $Y^{M}_\textsc{Seg}$ and $Y^{E}_\textsc{Seg}$ (e.g. in \Fref{fig:cs_target}).

It follows that the bilingual $p(Z|Z^M, Z^E)$ (Eq.~\eqref{eq:approx}) behaves as:
\begin{equation}
    z_t = 
    \begin{cases}
        m & \text{if $m \in \mathcal{V}^M \land e = \text{\textless}\textsc{Null}\text{\textgreater}$} \\
        e & \text{if $e \in \mathcal{V}^E \land m = \text{\textless}\textsc{Null}\text{\textgreater}$} \\
        b & \text{otherwise} \label{eq:2}
    \end{cases}
\end{equation}
where $m$ and $e$ are the arguments maximizing $p(z_t^{M}|X, z_{1:t-1}^M)$ and $p(z_t^{E}|X, z_{1:t-1}^E)$ respectively and $b$ is the argument maximizing $p(z_t|Z^M, Z^E, z_{1:t-1})$.
If either monolingual module predicts a CS point by emitting $\textless \textsc{Null} \textgreater$ then the bilingual module defaults to the prediction of the other monolingual module -- in other words, the first two cases of Eq.~\eqref{eq:2} expect that the language segmentation in Eq.~\eqref{eq:seg} is mistake-free.
The third fall-back case is considered for ambiguous language segmentation, such as if  $m$ and $e$ are both $\textless \textsc{Null} \textgreater$ or both non $\textless \textsc{Null} \textgreater$.
This case-by-case bilingual decision is an adverse design for our zero-shot objective -- models are likely to become over-reliant on the first two cases during training.
Language segmentation while training on purely monolingual utterances boils down to an over-simplified \textit{utterance-level} language identification task which may not generalize to \textit{intra-sententially} CS test utterances.
If CS point detection is expected to be tricky, then a more robust strategy should \textit{always} expect ambiguous monolingual inputs to the final bilingual decision as in the third case of Eq.~\eqref{eq:2}.

\section{Proposed Framework}
\label{sec:proposed}

In this section, we propose to completely remove language segmentation from monolingual modules using a transliteration-based formulation of $p(Z^{M/E}|X)$.
We then present a neural model of our modified conditionally factorized approach for zero-shot CS ASR.

\subsection{Modeling $p(Z^{M/E}|X)$ with Transliteration}
\label{sec:transliteration}

Rather than detecting CS points at the monolingual stage in order to know which speech segments to transcribe vs. which to ignore, we propose to simply allow each monolingual module to transcribe everything.
This means that for speech of a foreign language the monolingual modules are producing \textit{transliterations}, mapping sounds to phonetically similar units within their monolingual vocabularies $\mathcal{V}^M$ and $\mathcal{V}^E$. 
In other words, the monolingual modules simplify from Eq.~\eqref{eq:seg} to the following form (shown for Mandarin):
\begin{align}
    z_t^M = \argmax_{m\in \mathcal{V}^M \cup \{ \varnothing \}} p(z_t^M=m|X, z_{1:t-1}^M) \label{eq:trans}
\end{align}
where the speech $X$ may contain any language.
This form completely removes any sense of frame-wise language identity $\operatorname{LID}(\mathbf{x}_t)$.

To see why this modification is advantageous for zero-shot CS ASR, consider the corresponding change to the bilingual module:
\begin{equation}
    z_t = \argmax_{b \in  \mathcal{V}^M \cup \mathcal{V}^E \cup \{ \varnothing \}} p(z_t=b|Z^M, Z^E, z_{1:t-1}) \label{eq:new_bi}
\end{equation}
Note that this new bilingual form in Eq.~\eqref{eq:new_bi} never defaults to the prediction of one monolingual module as in the first two cases of the previously proposed bilingual form in Eq.~\eqref{eq:2}, reducing the risk of propagating errors made in the monolingual stage.
In other words, the bilingual decision now determines each $z_t$ by directly considering the conditional likelihood $p(z_t|Z^M, Z^E, z_{1:t-1})$ (Eq.~\eqref{eq:approx}). 
This modification effectively delays CS point detection from the monolingual stage (where we would have to simultaneously transcribe and perform frame-wise language identification per \Sref{sec:lang_seg}), to the bilingual stage (where transcription information is already given).

To train monolingual modules to transliterate speech segments of a foreign language, we obtain transliteration targets $Y^M_\textsc{Tra}$ and $Y^E_\textsc{Tra}$ using cross-lingual pseudo-labeling.\footnote{Unlike text-based transliteration \cite{knight1998machine}, pseudo-labeling relies solely on the resources presumed to be available in our zero-shot CS ASR settings.}
For instance, we pass monolingual English speech $X^M$ to a monolingual Mandarin ASR model $\textsc{ASR}^M(\cdot)$ for inference and vice versa as follows (e.g. in \Fref{fig:cs_target}):
\begin{align}
    Y^M_\textsc{Tra} \leftarrow \textsc{ASR}^M(X^E) \label{eq:7}\\
    Y^E_\textsc{Tra} \leftarrow \textsc{ASR}^E(X^M) \label{eq:8}
\end{align}
where $\textsc{ASR}^{M/E}(\cdot)$ denote generic label-to-frame models -- if we use the same architecture for pseudo-labeling as we do for our monolingual modules then these transliteration targets are cross-lingual semi-supervisions \cite{jyothi2015transcribing, thomas2020transliteration, billa21_interspeech, lugosch2022pseudo}.\footnote{We can apply transliteration to CS speech by stitching predictions corresponding to forced aligned \cite{kurzinger2020ctc} foreign segments between true native targets.}
Swapping the language segmentation targets $Y^M_\textsc{Seg}$ and $Y^E_\textsc{Seg}$ (\Sref{sec:lang_seg}) for these transliteration targets $Y^M_\textsc{Tra}$ and $Y^E_\textsc{Tra}$ is the \textit{only} modification required to realize our desired monolingual and bilingual module behaviors in Eq.~\eqref{eq:trans} and \eqref{eq:new_bi}.

\subsection{Conditional CTC with External LM Architecture}
\label{sec:conditional_ctc}

Finally, let us consider how to construct a neural architecture for our modified conditionally factorized framework.
Monolingual and bilingual label-to-frame posteriors (\Sref{sec:cond_background}) may be modeled using CTC or RNN-T networks as demonstrated by prior works \cite{cond2022, tian2022lae, song2022language}.
However for zero-shot CS ASR, the conditional independence assumption of CTC vs. the internal language modeling of RNN-T is a critical difference. 
A RNN-T based model may require internal language model (LM) adaptation \cite{rnnt_graves, meng2021internal, zhou2022language} to alleviate monolingual biases while a CTC based model can be directly applied to CS test sets with optional shallow external LM fusion \cite{hannun2014first}.

We therefore model monolingual, $p(Z^M|X)$ and $p(Z^E|X)$, and bilingual likelihoods, $p(Z|Z^M, Z^E)$, using CTC networks, $P_{\text{M\_CTC}}(\cdot)$, $P_{\text{E\_CTC}}(\cdot)$, and $P_{\text{B\_CTC}}(\cdot)$, as follows:
\begin{align}
    P_{\text{M\_CTC}}(z_t^M | X, \cancel{z_{1:t-1}^M}) &= \textsc{SoftmaxOut}^M(\mathbf{h}^{M}_{t}) \label{eq:9} \\
    P_{\text{E\_CTC}}(z_t^E | X, \cancel{z_{1:t-1}^E}) &= \textsc{SoftmaxOut}^E(\mathbf{h}^{E}_{t}) \label{eq:10} \\
    % \mathbf{h}^{B}_t &= \mathbf{h}_t^M + \mathbf{h}_t^E \label{eq:fuse} \\
    P_{\text{B\_CTC}}(z_t | \mathbf{h}^M, \mathbf{h}^E, \cancel{z_{1:t-1}}) &= \textsc{SoftmaxOut}^B(\mathbf{h}_t^M + \mathbf{h}_t^E) \label{eq:15}
\end{align}
where speech encoders, $\textsc{Encoder}^M$ and $\textsc{Encoder}^E$, map the speech signal, $X$, to latent monolingual representations, $\mathbf{h}^{M} = \{ \mathbf{h}_t^M \in \mathbb{R}^{D} | t = 1,  ... , T\}$ and $\mathbf{h}^{E} = \{ \mathbf{h}_t^E \in \mathbb{R}^{D} | t = 1,  ... , T\}$ followed by softmax normalized linear projections to monolingual or bilingual vocabularies.
Then addition fusion yields a bilingual latent representation which is finally fed to the bilingual CTC.
These three CTC networks are jointly optimized with an interpolated multi-task objective: $\mathcal{L} = \lambda_1 \mathcal{L}_\text{B\_CTC} + (1 - \lambda_1) (\mathcal{L}_\text{M\_CTC} + \mathcal{L}_\text{E\_CTC})/2$.

\begin{figure}
\centering
\includegraphics[width=\linewidth]{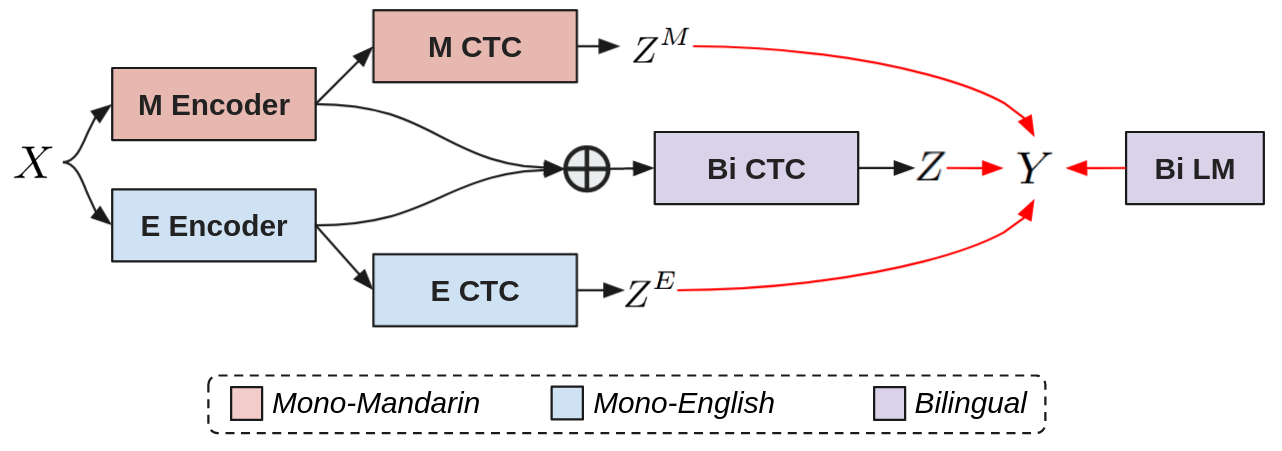}
\vspace{-5mm}
\caption{Conditional CTC architecture consisting of monolingual and bilingual CTC's plus an external bilingual LM. \textcolor{red}{Red} lines indicate joint decoding via time-synchronous beam search.}
\vspace{-2mm}
\label{fig:cond_ctc}
\end{figure}

During decoding, we first merge all CTC likelihoods, $P_{\text{M\_CTC}}(\cdot)$, $P_{\text{E\_CTC}}(\cdot)$, and $P_{\text{B\_CTC}}(\cdot)$, following the interpolation procedure described in Eq.~(6) of \cite{song2022language}; we denote this merged CTC likelihood as $P_{\text{CTC}}(Z|X)$.
We then jointly decode $P_{\text{CTC}}(\cdot)$ with an external bilingual LM, $P_{\text{B\_LM}}(Y)$, using the time-synchronous beam search described in \cite{hannun2014first}, which approximates the following decision:
\begin{align}
        \argmax_{Y\in \{\mathcal{V}^M \cup \mathcal{V}^E\}*} \lambda_2 ( \prod_{Z\in \mathcal{Z}} \log P_\text{CTC}(\cdot)) + (1-\lambda_2) \log P_\text{B\_LM}(\cdot) \label{eq:decoding}
\end{align}
where $\{\mathcal{V}^M \cup \mathcal{V}^E\}*$ denotes the set of all possible bilingual outputs.\footnote{For language segmentation variants of Conditional CTC, we do not expand hypotheses with the special $\textless \textsc{Null} \textgreater$ token to avoid corrupt outputs.}
This architecture, which we refer to as Conditional CTC, is depicted by the block-diagram in \Fref{fig:cond_ctc}.
The monolingual modules of these Conditional CTC models can perform either language segmentation (\Sref{sec:lang_seg}) or transliteration (\Sref{sec:transliteration}) depending on which set of monolingual targets (e.g. \Fref{fig:cs_target}) is used during training.
For transliteration, we obtain $Y^M_\textsc{Tra}$ and $Y^E_\textsc{Tra}$ (Eq.~\eqref{eq:7} and~\eqref{eq:8}) by greedily decoding monolingual CTC models (Eq.~\eqref{eq:9} and~\eqref{eq:10}) and then applying repeat and blank removal.

\section{Data and Experimental Setup}
% \vspace{-1mm}
\label{sec:setup}

\mypar{Data:} 
We apply $0.9$ and $1.1$ speed perturbations to up-sample SEAME \cite{lyu2010seame} training data by 3x, resulting in $291$h of total labeled speech data.
We then split the training data into CS and monolingual (Mandarin + English) parts to create two zero-shot settings.
The first setting allows $87$h of monolingual labeled speech data (for ASR training) and $89$k lines of unpaired CS or monolingual text data (for LM training).
The second fully zero-shot setting further removes the CS unpaired text data, leaving $39$k lines of unpaired monolingual text data.
Both settings remove $204$h of CS labeled speech data.
Monolingual CTC's trained on the English and Mandarin only SEAME splits were used for cross-lingual pseudo-labeling (\Sref{sec:transliteration}).

\mypar{Models:} Models are trained using ESPnet \cite{watanabe2018espnet}.
We combine $4000$ Mandarin characters with $4000$ English BPE \cite{sennrich2015neural} units to form the output vocabulary.
Conditional CTC models have $2$ conformer encoders \cite{gulati2020conformer, guo2021recent} with $12$ blocks, $4$ heads, $15$ kernel size, $2048$ feed-forward dim, $256$ and attention dim.
Vanilla CTC baselines with only $1$ encoder use $512$ attention dim, so all models have about $80$M parameters.
All models are initialized with encoder(s) pre-trained on $150$h of Mandarin AISHELL-1 \cite{bu2017aishell} and/or $118$h of English TED-LIUM-v1 \cite{rousseau-etal-2012-ted}.
We set $\lambda_1=0.7$ (\Sref{sec:conditional_ctc}) during training for $40$ epochs.
We set $\lambda_2=0.8$ (\Sref{sec:conditional_ctc}) during decoding with beam $10$.
We use RNN-LMs with $4$ layers and $2048$ dim trained for $20$ epochs.

\mypar{Evaluation:} Systems are evaluated on the full SEAME test sets (devman and devsge) and  also scored individually on the CS and monolingual portions of these sets.
We measure mixed error-rate (MER) that considers word-level English and character-level Mandarin.
% \vspace{-1mm}
\section{Results}
% \vspace{-1mm}
\label{sec:results}

\begin{table*}[t]
  \centering
    \caption{Results comparing Conditional CTC models with \textit{transliteration}-based monolingual modules to their \textit{language segmentation} counterparts and Vanilla CTC baselines. 
    The 1st horizontal partition shows top-line results when CS ASR training data is available. 
    The 2nd and 3rd partitions show zero-shot results when only monolingual ASR training data is available.
    Performances on the full, CS only, and monolingual only splits of the SEAME test sets are measured by \% mixed error rate (MER $\downarrow$). All models use CTC + LM decoding.
    }
    \resizebox {.92\linewidth} {!} {
\begin{tabular}{cll|cc|ccc|ccc}
\toprule
& & & ASR & LM & \multicolumn{3}{c|}{\underline{\textsc{DevMan}}} & \multicolumn{3}{c}{\underline{\textsc{DevSge}}}\\
\texttt{ID} & Model & Monolingual Behavior & Data & Data & Full & CS & M & Full & CS & M \\
\midrule
\texttt{A1} & Vanilla CTC \cite{hannun2014first} & \textit{No Monolingual Modules} & CS + M & CS + M & 18.8 & 18.2 & 21.5 & 26.2 & 23.7 & 29.8 \\
\texttt{A2} & Conditional CTC \cite{tian2022lae, song2022language} & Language Segmentation & CS + M & CS + M & \underline{\textbf{17.1}} & \textbf{16.5} & 19.9 & \underline{\textbf{23.5}} & \textbf{21.4} & \textbf{26.5} \\
\texttt{A3} & Conditional CTC (Ours) & Transliteration & CS + M & CS + M & 17.3 & 16.9 & \textbf{19.1} & 24.0 & 22.1 & 26.7 \\
\midrule
\texttt{B1} & Vanilla CTC \cite{hannun2014first} & \textit{No Monolingual Modules} & M & CS + M & 36.6 & 38.9 & 27.0 & 42.5 & 47.0 & 36.1 \\
\texttt{B2} & Conditional CTC \cite{tian2022lae, song2022language} & Language Segmentation & M & CS + M & 30.1 & 32.0 & 22.0 & 35.7 & 39.7 & \textbf{30.1} \\
\texttt{B3} & Conditional CTC (Ours) & Transliteration & M & CS + M & \underline{\textbf{25.2}} & \textbf{26.0} & \textbf{21.9} & \underline{\textbf{31.0}} & \textbf{31.5} & 30.2 \\
\midrule
\texttt{C1} & Vanilla CTC \cite{hannun2014first} & \textit{No Monolingual Modules} & M & M & 39.1 & 41.6 & 28.4 & 44.8 & 50.0 & 37.3 \\
\texttt{C2} & Conditional CTC \cite{tian2022lae, song2022language} & Language Segmentation & M & M & 32.2 & 34.4 & 23.0 & 37.8 & 42.6 & 31.1 \\
\texttt{C3} & Conditional CTC (Ours) & Transliteration & M & M & \underline{\textbf{27.3}} & \textbf{28.5} & \textbf{22.6} & \underline{\textbf{32.7}} & \textbf{34.0} & \textbf{30.8}\\
\bottomrule
\end{tabular}
}
    \label{tab:main}
    % \vspace{-3mm}
\end{table*}

\Tref{tab:main} presents results in three horizontal partitions where 1) all SEAME training data is allowed 2) CS speech data is removed and 3) CS speech and text data are removed; the latter two settings emulate practical zero-shot scenarios.
When CS speech data is available, language segmentation is reliable and thus the transliteration-based method is not necessary (\texttt{A2} vs. \texttt{A3}).
However, once CS speech data is removed the language segmentation approach degrades $13$ absolute MER on both full test sets; as a result the transliteration approach outperforms by $5$ absolute MER, a wide margin, owing primarily to superior performance on CS utterances (\texttt{B2} vs. \texttt{B3}).
When CS text data is also removed both variants of Conditional CTC degrade only by an additional $2$ absolute MER and the gap between remains (\texttt{C2} vs. \texttt{C3}).
In all three data settings both Conditional CTC models outperform Vanilla CTC baselines.

\subsection{Ablations on the Conditional CTC Model}

\begin{table}[t]
    \vspace{-3mm}
  \centering
    \caption{Ablation study examining the relative importance of monolingual CTC, bilingual CTC, and bilingual LM modules during decoding as measured by \% mixed error rate (MER $\downarrow$) on the devman test set. Bilingual modules are shown in \textcolor{blue}{blue} and the most severely degraded combination (with no bilingual modules) is \textbf{bolded}.}
    \resizebox {\linewidth} {!} {
\begin{tabular}{cll|c}
\toprule
\texttt{\#} & Model & Decoding Likelihoods & MER($\downarrow$)\\
\midrule
\texttt{1} & Cond. CTC w/ Trans. & $P_\text{M\_CTC}$, $P_\text{E\_CTC}$, \textcolor{blue}{$P_\text{B\_CTC}$}, \textcolor{blue}{$P_\text{B\_LM}$} & 25.2 \\
\texttt{2} & $-$ Bilingual LM & $P_\text{M\_CTC}$, $P_\text{E\_CTC}$, \textcolor{blue}{$P_\text{B\_CTC}$} & 27.4 \\
\texttt{3} & $-$ Monolingual CTCs & \textcolor{blue}{$P_\text{B\_CTC}$}, \textcolor{blue}{$P_\text{B\_LM}$} & 25.7 \\
\texttt{4} & \hspace{1.5em}$-$ Bilingual LM & \textcolor{blue}{$P_\text{B\_CTC}$} & 27.9 \\
\texttt{5} & $-$ Bilingual CTC & $P_\text{M\_CTC}$, $P_\text{E\_CTC}$, \textcolor{blue}{$P_\text{B\_LM}$} & 26.0 \\
\texttt{6} & \hspace{1.5em}$-$ Bilingual LM & $P_\text{M\_CTC}$, $P_\text{E\_CTC}$ & \textbf{48.1} \\
\bottomrule
\end{tabular}
}

    \label{tab:ablation}
    % \vspace{-2mm}
\end{table}

Our Conditional CTC models consist of three types of modules: monolingual CTC's ($P_{\text{M\_CTC}}$ and $P_{\text{E\_CTC}}$), bilingual CTC ($P_{\text{B\_CTC}}$), and bilingual LM ($P_{\text{B\_LM}}$).
In \Tref{tab:ablation}, we examine the relative contributions of these modules by removing each from model $\texttt{B3}$ of \Tref{tab:main} during joint decoding (described in \Sref{sec:conditional_ctc}).
Removing the bilingual LM (line \texttt{2}) degrades performance more than removing the bilingual CTC (line \texttt{5}), showing the importance of utilizing CS textual data when available.
Further, note that monolingual CTCs do contribute (line \texttt{3}), but are insufficient on their own (line \texttt{6}).
Finally, the fact performance is still reasonable without the bilingual CTC (line \texttt{5}) suggests that separately trained monolingual CTCs may be directly applied to CS ASR if a CS LM is available -- this direction may offer a high degree of scalability towards the long-tail of possible CS pairs and towards CS between three or more languages.

\subsection{Relaxing the Zero-Shot Setting}

\begin{figure}
\centering
\includegraphics[width=.8\linewidth]{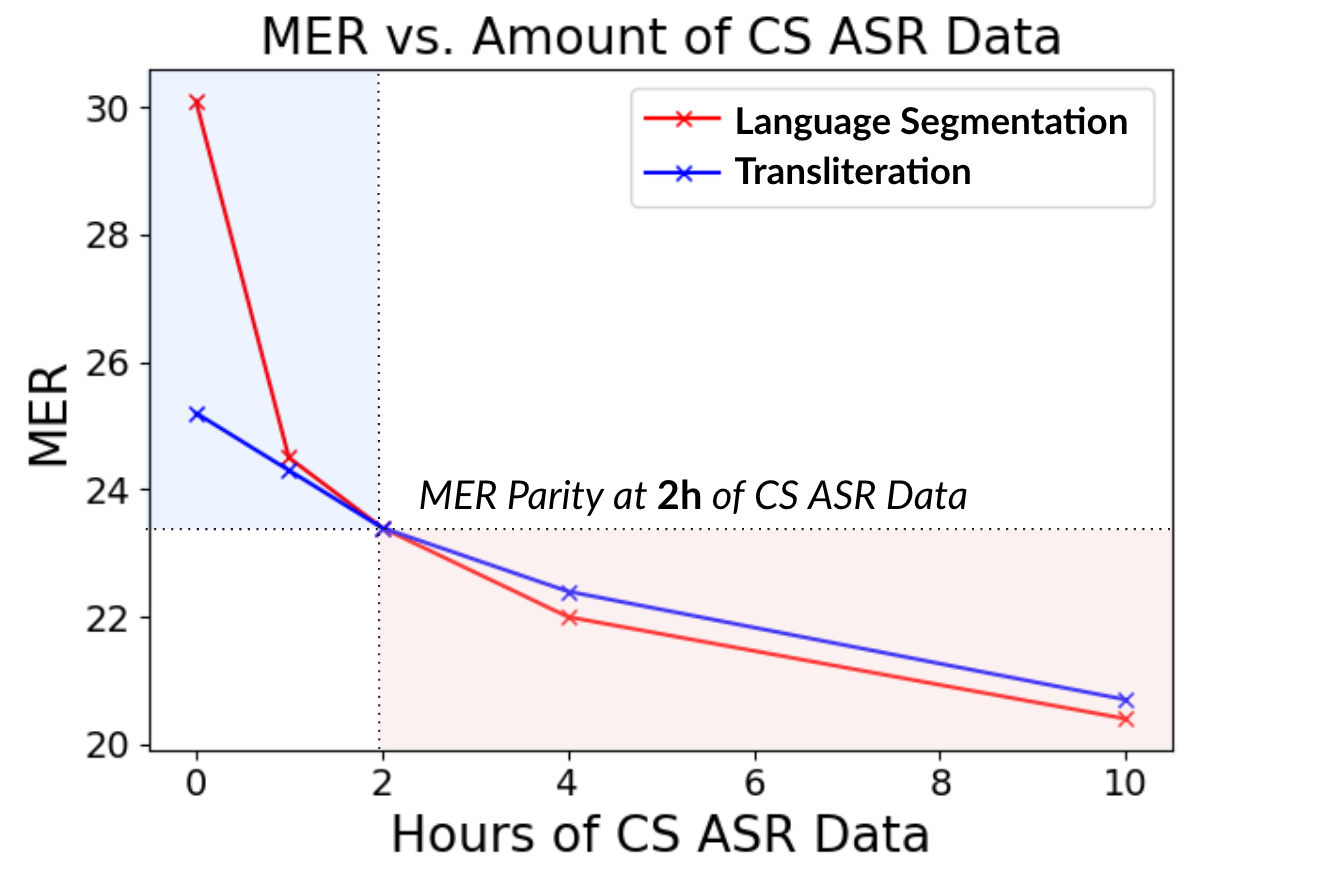}
\vspace{-2mm}
\caption{Analysis on the amount of CS ASR training data required for conditional CTC with language segmentation to outperform conditional CTC with transliteration. MER($\downarrow$) on devman is shown.}
\label{fig:data_study}
\vspace{-2mm}
\end{figure}

How much CS ASR training data do we need for the originally proposed language segmentation method (\Sref{sec:lang_seg}) to be sufficient?
The answer depends on the proximity of the particular language pair and characteristics of the dataset being used, but in our experimental setup we find that the answer is $2$h of CS speech data (see \Fref{fig:data_study}).
The decreasing effectiveness of our transliteration method for increasing amounts of CS ASR training data suggests that the cross-lingual pseudo-labels are noisy to a degree.
Future investigations into improving pseudo-labeling quality (e.g. via constrained decoding) may benefit this work and other related techniques which employ cross-lingual semi-supervision \cite{jyothi2015transcribing, thomas2020transliteration, billa21_interspeech, lugosch2022pseudo}.

% \vspace{-1mm}
\section{Conclusion}
\vspace{-0.5mm}

We identify that the promising conditionally factorized joint CS and monolingual ASR framework has an acute weakness which limits its applicability to zero-shot CS ASR; the original formulation expects that each monolingual module can cleanly transcribe native speech while ignoring foreign speech.
We propose a simple modification via cross-lingual pseudo-labeling to allow the monolingual modules to instead produce transliterations of foreign speech, thereby avoiding error propagation of frame-wise LID decisions.
We demonstrate the effectiveness of our transliteration-based method using Conditional CTC models deployed for zero-shot Mandarin-English CS ASR.
In future work, we will extend to other languages, scale beyond bilingualism, and refine our pseudo-labeling technique.
\blfootnote{This work was supported by JSALT 2022 at JHU via Amazon, Microsoft and Google. Brian Yan, Matthew Wiesner, and Shinji Watanabe are also supported by the Human Language Technology Center of Excellence at JHU.}

\vfill\pagebreak

\section{References}
{
\printbibliography
}

\end{document}